\def\year{2022}\relax
\documentclass[letterpaper]{article} % DO NOT CHANGE THIS
\usepackage{aaai22}  % DO NOT CHANGE THIS
\usepackage{times}  % DO NOT CHANGE THIS
\usepackage{helvet}  % DO NOT CHANGE THIS
\usepackage{courier}  % DO NOT CHANGE THIS
\usepackage[hyphens]{url}  % DO NOT CHANGE THIS
\usepackage{graphicx} % DO NOT CHANGE THIS
\usepackage{natbib}  % DO NOT CHANGE THIS AND DO NOT ADD ANY OPTIONS TO IT
\usepackage{caption} % DO NOT CHANGE THIS AND DO NOT ADD ANY OPTIONS TO IT
\DeclareCaptionStyle{ruled}{labelfont=normalfont,labelsep=colon,strut=off} % DO NOT CHANGE THIS
\usepackage{algorithm}
\usepackage{algorithmic}

\usepackage{newfloat}
\usepackage{listings}
\floatstyle{ruled}
\newfloat{listing}{tb}{lst}{}
\floatname{listing}{Listing}
\title{Combining Deep Learning and Reasoning for Address Detection in Unstructured Text Documents}
\author{
   Matthias Engelbach\textsuperscript{\rm 1}, Dennis Klau\textsuperscript{\rm 2}, Jens Drawehn\textsuperscript{\rm 1}, Maximilien Kintz\textsuperscript{\rm 1}\\
}
\affiliations{
      \textsuperscript{\rm 1}Fraunhofer Institute for Industrial Engineering IAO\\Nobelstr. 12\\70569 Stuttgart,  Germany\\
			\textsuperscript{\rm 2}University of Stuttgart, Institute of Human Factors and Technology Management IAT\\Nobelstr. 12\\70569 Stuttgart, Germany\\
     \{matthias.engelbach, jens.drawehn, maximilien.kintz\}@iao.fraunhofer.de, dennis.klau@iat.uni-stuttgart.de
\vspace*{10pt}
}

\usepackage{bibentry}
\usepackage{array}

\begin{document}

\maketitle

\begin{abstract}
Extracting information from unstructured text documents is a demanding task, since these documents can have a broad variety of different layouts and a non-trivial reading order, like it is the case for multi-column documents or nested tables. Additionally, many business documents are received in paper form, meaning that the textual contents need to be digitized before further analysis. Nonetheless, automatic detection and capturing of crucial document information like the sender address would boost many companies' processing efficiency. 
In this work we propose a hybrid approach that combines deep learning with reasoning for finding and extracting addresses from unstructured text documents. We use a visual deep learning model to detect the boundaries of possible address regions on the scanned document images and validate these results by analyzing the containing text using domain knowledge represented as a rule based system. 
\end{abstract}

\section{Introduction}

Many businesses need to deal with scanned text documents, such as invoices or purchase orders, on a daily basis. This continuing trend is reflected by the importance of OCR solutions worldwide \cite{ocr_importance}. Nowadays, many powerful tools like OCR engines or deep learning models, trained on large quantities of data, are freely available and can be used for the automated analysis of such documents. However, adoption of these generic tools for specific tasks is still error-prone, since human expert domain knowledge is required in many use cases. This \textit{a priori} domain knowledge is usually not well represented or integrated in modern AI models \cite{aiforscience}. 
For instance, before continuing to process the contents of a document, we found that a first crucial step for businesses is usually a proper identification of the correct meta data, like sender and receiver address, because this information might influence the subsequent processing flow. Humans can solve this task easily, since they know about typical locations of address information on document headers as well as the internal structure of addresses that all entities have to follow. Our experience showed, that pure machine learning solutions can heavily profit from such human reasoning assessments, provided these information can be represented and integrated in a learning system. 

For tackling this problem of address detection, we built a processing pipeline, consisting of different data-driven and rule-based components, which was developed in cooperation with the University of Leipzig. Many of the components originally come from the open source OCR-D project~\cite{OCRD}, which aims to provide means for analysis of scanned documents by enabling usage of other (heterogeneous) open source processing tools, like e.g. Tesseract OCR \cite{tesseract}, in a standardized way.

Our work is structured as follows: We first give a short overview of related work, followed by a comparison of data-driven and reasoning-based approaches in general. This forms the basis for further technology decisions during the pipeline implementation described in the subsequent section. Thereafter, we describe the evaluation of our approach using a document test set and conclude our work with an outlook and future work.

\begin{table*}[h!]
	\centering
	\small
	\begin{tabular}{|p{4cm}|p{4cm}|p{4cm}|p{4cm}|}
		\hline
		\textbf{Criteria} & \textbf{Deep Learning} & \textbf{Reasoning} & \textbf{Consequence}\\
		\hline
		\emph{Availability of training data or example documents} & Quantity of training data limited, especially in small businesses, leading to lacking model quality & Number of sample documents typically sufficient for the definition of simple rules & Risk that deep learning models generalize poorly (and produce too many false negatives) or generalize too much (and produce too many false positives) \\
		\hline
		\emph{Availability of reference databases} (for example Customer Relationship Management system) & Typically ignored by standard deep learning approaches & Useful for implementing plausibility checks / validation rules & Integration of validation rules using reference database in a hybrid approach to reduce false positives \\
		\hline
		\emph{Use of standard layout for internal documents} & Result in usually high-quality models, even with limited training data & Well-suited for writing visual rules & Possibility to use visual rules for processing of internal documents\\
		\hline
		\emph{High variability of layouts for external documents} (i.e., from customers) & - & Increases the difficulty of defining rules & Need for deep learning models tackling the variety of designs \\
		\hline
		\emph{False positives} (wrong address being extracted) lead to faulty document processing (e.g., routing to the wrong department or person) & Tend to be common & Tend to be rare when using validation rules & Need to limit false positives by combining reasoning and validation after deep learning extraction \\
		\hline
		\emph{False negatives} (no address being extracted) lead to the need for manual data extraction & Tend to be rarer than with reasoning approaches & Tend to be common as rules generalize poorly & False negatives can be accepted, but they limit the possible degree of full automation \\
		\hline
		\emph{Debugging} & Black-box character of deep learning models and limitations of XAI usually only allow the use of additional training data with debugging and improvement of model quality & Software developers typically can fine-tune the rules when provided with sample documents that were incorrectly processed & No significant difference between approaches as business customers tend to prefer stable and slowly evolving solutions\\
		\hline
	\end{tabular}
	\caption{Comparison of deep learning and reasoning approaches for address data extraction from German-language business correspondence}
	\label{tab:comparison}
\end{table*}

\section{Related Work}
\label{sec:related}
State-of-the-art document analysis models, like the multi-modal LayoutLM architecture released by Microsoft, target this problem by combining both visual and text information in the model architecture, even for multiple languages~\cite{YihengXu.2021}. Nonetheless, results achieved by pure data driven approaches always retain some amount of uncertainty regarding plausibility and validity from the view of a human expert. Therefore, we propose an approach that merges machine intelligence and human reasoning for improved address detection.

Extracting address information from unstructured texts is not a new use case: Long before the rise of modern AI and deep learning technology, rule-based approaches and algorithms aimed to detect post addresses in digitized texts, for example on web pages~\cite{Can.2005}. With the upcoming of machine learning techniques, more data driven approaches were applied, for example using Conditional Random Fields~\cite{Chang.2010}. However, when dealing with complex document layouts or noisy scanned documents, the correct identification and boundary detection in terms of entity extraction remains error-prone. 

A similar approach to our pipeline was developed by \cite{deepreader}, where an enterprise-based platform was built that attempts to populate a general relational hierarchy using document templates prior to information extraction and then mapping the found information and hierarchy into a database. While the authors and our work share similar preprocessing approaches like noise-reduction techniques or information retrieval using OCR, we rely on capturing the relational and document structure information implicitly by a combination of prior domain knowledge or business rules and the deep learning model without templates. 
The authors of \cite{neuro_synth} propose a two-folded approach for finding specific text entities. In the first step, they use a pre-trained model to generate a relational database similar to \cite{deepreader} and afterwards apply deductive reasoning to learn extraction programs.

In this paper, we focus more on specific use-cases for information localization and extraction, as well as validation of ambiguous entities by extracting structural information without predefined templates using deep neural networks and incorporating prior domain knowledge or business rules, in the aforementioned document classes.

\begin{figure*}[h!]
	\centering
	\includegraphics[width=1.0\textwidth]{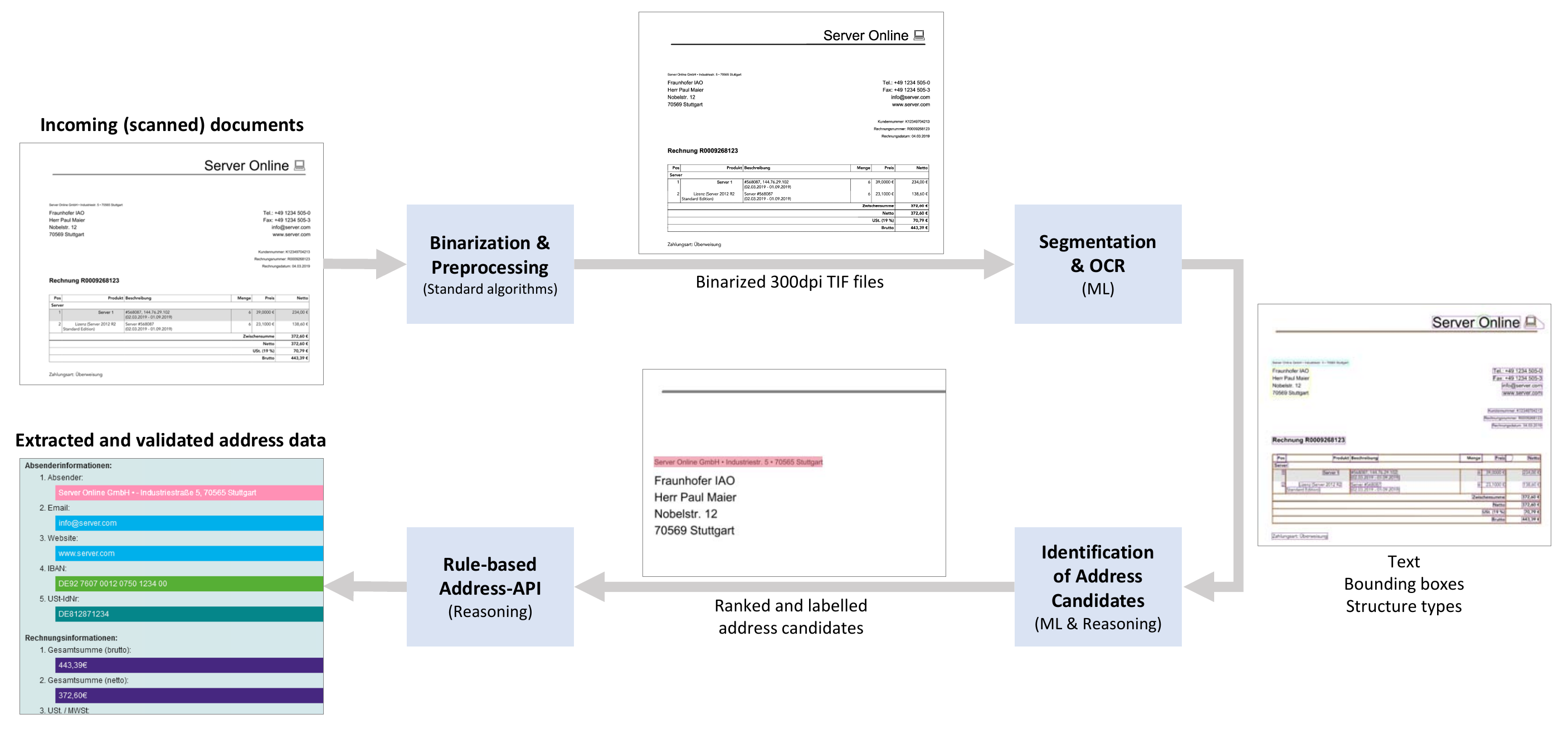}
	\caption{Processing pipeline for address detection in business documents}
	\label{fig:process}
\end{figure*}

\section{Comparison of Deep Learning and Reasoning for Address Extraction}
\label{sec:comparison}

Our work focuses on the extraction of sender or receiver address data in business correspondence, as they appear in invoices, purchase orders or customer letters. This use case was chosen based on the feedback of many project partners and customers: Extracting address data and matching the sender or receiver of documents with a database is very often the first step in automated document processing. The restriction to business correspondence implies some specificity that shows the strengths and limits of sole deep learning and reasoning approaches listed in Table~\ref{tab:comparison}, which led to the hybrid approach showcased in this work.

Recent work has shown that approaches such as transfer learning may reduce the downsides of deep learning when little training data is available~\cite{Martinek.2020}. The application of transfer learning, however, requires the availability of sufficiently good foundational models~\cite{bommasani2021opportunities}. These may be available for historical -- mostly anglophone -- business correspondence where training data is available, like the RVL-CDIP data set \cite{AdamWHarley.2015}. To the best of our knowledge, however, no sufficient publicly available datasets exists for business correspondence in German.
Overall, the comparison of both pure methods indicates, that a hybrid approach combining the generalization capability of deep learning to a variety of layouts and the reduction of false positives in typical reasoning, appears promising.

\section{Hybrid Method for Address Extraction}
\label{sec:method}

\subsection{Problem Definition}

In our use case, the goal of address detection is achieved in two steps:
\begin{enumerate}
	\item Detection of possible address candidates on the (scanned) pages of business documents
	\item Assignment of the correct address class for each candidate (\emph{sender}, \emph{receiver}, \emph{other})
\end{enumerate}

\noindent Note that the found \emph{sender} and \emph{receiver} address are limited to one instance per document page, while the amount of \emph{other} addresses remains unlimited. An address candidate is considered as detected correctly, if the text within its region boundaries contains at least the ZIP Code and the city name of the correct address.

\subsection{Pipeline Realization}

\begin{table*}[]
	\centering
	\small
	\begin{tabular}{|p{4cm}|p{4cm}|p{4cm}|p{4cm}|}
		\hline
		\textbf{Pipeline Component} & \textbf{Deep Learning} & \textbf{Reasoning} & \textbf{Consequence / Technology Choice}\\
		\hline
		\emph{Binarization} & - & Standard image processing algorithms & Standard algorithms are sufficient here\\
		\hline
		\emph{Segmentation} & For complex and diverse layouts suitable & Rules would quickly become too complex & Usage of pre trained data driven models \\
		\hline
		\emph{OCR} & Nowadays, best results are achieved here with deep learning & Not used anymore & Usage of Tesseract 4.0 working based on deep learning approach\\
		\hline
		\emph{Address Region Detection} & For complex and diverse layouts suitable & Rules would quickly become too complex & Visual Deep Learning approach was chosen. \\
		\hline
		\emph{Address Text Detection} & Customized Named Entity Recognition Models thinkable, but training data is needed & Advantages of the use of reference databases and plausibility checks & Usage of rule based validation because of well defined structure of German addresses  \\
		\hline
	\end{tabular}
	\caption{Description of the influence and application of data-driven and rule-based methods in the respective pipeline steps}
	\label{tab:choice}
\end{table*}

We built a prototype in form of the processing pipeline which utilizes a hybrid workflow as is shown in Figure~\ref{fig:process}, including both: deep learning and rule-based pattern matching. The choice of technology for the individual pipeline steps was based on the considerations listed in Table \ref{tab:choice}. 

In the following, we describe the realization of different processing steps in the pipeline. For many of the components, modules from the OCR-D project were adapted to our specific use case of address detection\footnote{for details to the individual components see https://github.com/OCR-D}.

In general, the pipeline is working on scanned documents like invoice or delivery documents or any other type of correspondence letters that might arrive in the form of PDFs without text layer or any image format. At the pipeline entry, all incoming documents are converted to 300dpi TIF files and the pages of each document are segregated. In the following step, the file is binarized with the tool Olena, included in the OCR-D project. 

Afterwards, a general visual document segmentation is performed using Tesseract and OCRopy, an open-source document analysis framework which allows identification of table regions and structures~\cite{ocropy}. As a result of this step, we receive all text regions and lines that have been identified, together with their bounding box coordinates on the page. The OCR of Tesseract 4.0 is applied for extracting the textual contents of the detected lines.

For identification of address region candidates, we send the OCR results to our reasoning-based Address-API component. The module recognizes address-like sections using human crafted rules. Specifically, two steps of processing are performed:
\begin{enumerate}
	\item \textbf{Detection of address components}: The module detects individual address components like ZIP codes, cities, streets or names of persons and organizations by using regular expressions and reference lists. For instance, ZIP codes are found using a regular term matching all string tokens containing exactly 5 digits. These candidates are then examined using a list of valid German ZIP Codes and discarded in the following steps if they are found to be invalid.
	\item \textbf{Composition of complex address entities}: The found base entities from the previous step are composed to complex address entities according to the rules of German address structures, usually following this basic scheme: [ADRESSEE (person or organization)] [STREET and HOUSENUMBER] [ZIP CODE and CITY]. This way, a prediction can be made, whether a given section is a proper an address. In detail, a confidence score is computed depending on the individual validation scores of the sub entities and their plausibility in the whole address context. For instance, ZIP codes are matched against city names found close to them by lookups in the German ZIP-city registry.
\end{enumerate}

Finally, the address prediction for each text line is combined with the visual information, like region image, boundaries and location on the document page. This is done by feeding both visual and text classification information into a modified Deep Learning model for predicting the final address region boundaries together with their address label (\emph{sender}, \emph{receiver}, \emph{other}). More precisely, we fine-tuned a Mask R-CNN model~\cite{matterport_maskrcnn_2017} using both synthetically generated letters (with corresponding address labels) and about 300 annotated internal invoice documents of different layouts and scan quality. Since these documents have been taken from our internal ordering processes, the often also contain additional noise like small portions of handwritten notes or company stamps that have been put on the pages, also sometimes covering some of the original text data.

At the end of the pipeline, the textual contents of the predicted address regions for \emph{sender}, \emph{receiver} and \emph{other} are again validated by the same reasoning-based approach already used in the previous step (Address-API) to make sure we really received a valid address region from the model. This also includes utilization of a geocoding service for normalizing and validating detect address information.
Note that although the conception and implementation of these pipeline steps have mostly been done, our approach described here is still work in progress. Hence, the following evaluation is still based on the results predicted by the deep learning model mentioned in the last step and thus constituting only a first step towards indicating the potential of our hybrid approach for address detection. We expect the results to further improve by introducing this second level of (repeated) textual validation in our future work.

%\begin{enumerate}
	%\item preprocessing with LW (binarization, cropping, other OCR-D process steps)
	%\item detection of potential address regions with visual deep learning model (Mask-RCNN)
	%\item sending textual contents to address API
	%\item Merging results of visual and textual classifications for determining final address regions.
%\end{enumerate}

\section{Evaluation}
\label{sec:evaluation}

\subsection{Test Data}
For evaluating the quality of our address extraction approach, we built a test set consisting of 64 scanned documents with a resolution of 300 dpi. The data set provides a total of 104 pages, of which 67 pages contain a \emph{sender} and 71 a \emph{receiver} address. Additionally, 105 addresses of type \emph{other} are distributed across 74 pages. There are 8 pages left without any addresses. 

\hspace*{5pt}
\begin{center}
	\begin{tabular}{| >{\raggedright\arraybackslash}p{1.5cm} | >{\raggedleft\arraybackslash}p{1cm} | >{\raggedleft\arraybackslash}p{1cm} | >{\raggedleft\arraybackslash}p{1cm} | >{\raggedleft\arraybackslash}p{1cm} |}
		\hline
		\phantom{0} & \textbf{TP} & \textbf{FP} & \textbf{FN} & {\raggedleft\textbf{F1}}\\ 
		\hline
		\emph{Sender} & 39 & 0 & 27 & \textbf{0.7429}\\
		\hline
		\emph{Receiver} & 57 & 0 & 13 & \textbf{0.8976}\\
		\hline
		\emph{Other} & 41 & 15 & 64 & \textbf{0.5093} \\
		\hline
		\emph{All} & 137 & 15 & 104 & \textbf{0.6972}\\
		\hline
	\end{tabular}
	\captionof{table}{Evaluation results on test set using the hybrid address extraction methodology}
	\label{tab:result}
\end{center}

\subsection{Results and Discussion}

%\hspace*{10pt}
In our experiments, we consider an address as correctly detected and classified (TP) if the predicted region contains at least the core components of an address (postal code and city, and additionally street and house number if present in the ground truth) and is assigned the correct class label (\emph{sender}, \emph{receiver}, \emph{other}). Regions that were predicted as addresses and contain no address information are defined as false positives (FP). Existing address regions in the test set that were not recognized or got the wrong address type assigned are false negatives (FN). The performance results on the whole test set are listed in Table~\ref{tab:result}. Note that we leave out the indication of true negatives (TN), since we would have to consider all other non-address text regions that were correctly not labeled as addresses as well.

The low occurrence of FP indicates that our hybrid approach is able to predict address regions with a high degree of reliability, which is due to our reasoning-based textual validation mechanism, whose result is directly used as secondary input to the visual deep learning model. The high amount of false negatives shows that we are missing a lot of addresses. In many cases, we were able to trace back the reason for this to insufficient quality of the OCR results, which leads to incorrect address validation predictions by the reasoning-based component. This is due to the strict address formatting rules enforced by the module. Hence, if for example single characters in either the postal code, city or street name are incorrectly extracted by the OCR process, the match to our reference list will fail with high probability. In future works, we aim to tackle this problem by improving the OCR component and allowing more fuzzy matching in the reasoning and rule-based parts of the pipeline.

\section{Conclusion and Future Work}
\label{sec:conclusion}
Our experiments have shown that combining the strengths of data-driven and reasoning-based approaches for the use case of address extraction (in form of domain knowledge about structure and contents of country-specific addresses) can improve stability and reliability of results. For our future work, we plan to optimize each step of the pipeline. For the deep learning models, this means acquiring more training and fine tuning on bigger data sets. For the reasoning-based parts, this implies to soften the strictness of rule validation by allowing a small amount of fuzziness. Additionally, we aim to extend our approach to other tasks of information extraction, in particular intelligent analysis of complex or nested table structures. Finally, we are planning to perform bigger automated evaluations on the whole pipeline results as well as on the single outcomes of each pipeline component for measuring progress and success of our approach.

\appendix

\label{sec:references}

% Use \bibliography{yourbibfile} instead or the References section will not appear in your paper
\bibliography{aaai22}

\begin{thebibliography}{14}
\providecommand{\natexlab}[1]{#1}

\bibitem[{Abdulla(2017)}]{matterport_maskrcnn_2017}
Abdulla, W. 2017.
\newblock Mask R-CNN for object detection and instance segmentation on Keras
  and TensorFlow.
\newblock \url{https://github.com/matterport/Mask_RCNN}.

\bibitem[{Bommasani et~al.(2021)Bommasani, Hudson, Adeli, Altman, Arora, von
  Arx, Bernstein, Bohg, Bosselut, Brunskill
  et~al.}]{bommasani2021opportunities}
Bommasani, R.; Hudson, D.~A.; Adeli, E.; Altman, R.; Arora, S.; von Arx, S.;
  Bernstein, M.~S.; Bohg, J.; Bosselut, A.; Brunskill, E.; et~al. 2021.
\newblock On the opportunities and risks of foundation models.
\newblock Stanford University: Center for Research on Foundation Models (CRFM),
  Stanford Institute for Human-Centered Artificial Intelligence (HAI).

\bibitem[{Breuel(2017)}]{ocropy}
Breuel, T. 2017.
\newblock OCRopus: document analysis and optical character recognition (OCR)
  system.
\newblock \url{https://github.com/ocropus/ocropy}.
\newblock Accessed: 2021-11-10.

\bibitem[{Chang and Li(2010)}]{Chang.2010}
Chang, C.-H.; and Li, S.-Y. 2010.
\newblock MapMarker: Extraction of Postal Addresses and Associated Information
  for General Web Pages.
\newblock volume~1, 105--111.

\bibitem[{{Grand View Research Inc.}(2019)}]{ocr_importance}
{Grand View Research Inc.} 2019.
\newblock Optical Character Recognition Market Size Report, 2021-2028.
\newblock
  \url{https://www.grandviewresearch.com/industry-analysis/optical-character-recognition-market}.
\newblock Accessed: 2021-11-10.

\bibitem[{Harley, Ufkes, and Derpanis(2015)}]{AdamWHarley.2015}
Harley, A.~W.; Ufkes, A.; and Derpanis, K.~G. 2015.
\newblock Evaluation of Deep Convolutional Nets for Document Image
  Classification and Retrieval.
\newblock In \emph{International Conference on Document Analysis and
  Recognition (ICDAR)}.

\bibitem[{Lin et~al.(2005)Lin, Zhang, Meng, and Lin}]{Can.2005}
Lin, C.; Zhang, Q.; Meng, X.; and Lin, W. 2005.
\newblock Postal Address Detection from Web Documents.
\newblock In \emph{2005 International Workshop on Challenges in Web Information
  Retrieval and Integration {(WIRI} 2005), 8-9 April 2005, Tokyo, Japan},
  40--45. {IEEE} Computer Society.

\bibitem[{Mart{\'i}nek, Lenc, and Kr{\'a}l(2020)}]{Martinek.2020}
Mart{\'i}nek, J.; Lenc, L.; and Kr{\'a}l, P. 2020.
\newblock Building an efficient OCR system for historical documents with little
  training data.
\newblock \emph{Neural Computing and Applications}, 32(23): 17209--17227.

\bibitem[{Neudecker et~al.(2019)Neudecker, Baierer, Federbusch, Boenig,
  W\"{u}rzner, Hartmann, and Herrmann}]{OCRD}
Neudecker, C.; Baierer, K.; Federbusch, M.; Boenig, M.; W\"{u}rzner, K.-M.;
  Hartmann, V.; and Herrmann, E. 2019.
\newblock OCR-D: An End-to-End Open Source OCR Framework for Historical Printed
  Documents.
\newblock In \emph{Proceedings of the 3rd International Conference on Digital
  Access to Textual Cultural Heritage}, DATeCH2019, 53–58. New York, NY, USA:
  Association for Computing Machinery.

\bibitem[{Smith(2019)}]{tesseract}
Smith, R. 2019.
\newblock Tesseract OCR: an optical character recognition engine for various
  operating systems.
\newblock \url{https://github.com/tesseract-ocr/tesseract}.
\newblock Accessed: 2021-11-10.

\bibitem[{Stevens et~al.(2020)Stevens, Taylor, Nichols, Maccabe, Yelick, and
  Brown}]{aiforscience}
Stevens, R.; Taylor, V.; Nichols, J.; Maccabe, A.~B.; Yelick, K.; and Brown, D.
  2020.
\newblock AI for science, Report on the Department of Energy (DOE) Town Halls
  on Artificial Intelligence (AI) for Science.

\bibitem[{Sunder et~al.(2019)Sunder, Srinivasan, Vig, Shroff, and
  Rahul}]{neuro_synth}
Sunder, V.; Srinivasan, A.; Vig, L.; Shroff, G.; and Rahul, R. 2019.
\newblock One-shot Information Extraction from Document Images using
  Neuro-Deductive Program Synthesis.
\newblock In \emph{13th International Workshop on Neural-Symbolic Learning and
  Reasoning}, 11.

\bibitem[{Vishwanath et~al.(2019)Vishwanath, Rahul, Sehgal, ~, Chowdhury,
  Sharma, Vig, Shroff, and Srinivasan}]{deepreader}
Vishwanath, D.; Rahul, R.; Sehgal, G.; ~, S.; Chowdhury, A.; Sharma, M.; Vig,
  L.; Shroff, G.; and Srinivasan, A. 2019.
\newblock Deep Reader: Information Extraction from Document Images via Relation
  Extraction and Natural Language.
\newblock In \emph{Computer Vision – ACCV 2018 Workshops}, 186--201. Springer
  International Publishing.
\newblock ISBN 978-3-030-21073-1.

\bibitem[{Xu et~al.(2021)Xu, Lv, Cui, Wang, Lu, Florencio, Zhang, and
  Wei}]{YihengXu.2021}
Xu, Y.; Lv, T.; Cui, L.; Wang, G.; Lu, Y.; Florencio, D.; Zhang, C.; and Wei,
  F. 2021.
\newblock LayoutXLM: Multimodal Pre-training for Multilingual Visually-rich
  Document Understanding.

\end{thebibliography}

\end{document}